# Enhancing Scientific Named Entity Recognition via Large Language Models: A Type-driven Multi-task Learning Approach

Tong Bao, Yi Zhao, Heng Zhang, Chengzhi Zhang*
Department of Information Management, Nanjing University of Science and Technology, Nanjing, 210094, China

**Abstract**: Scientific named entity recognition (SciNER) plays a crucial role in information extraction and knowledge discovery from scientific texts. Recently, large language models (LLMs) have demonstrated the capacity to achieve competitive SciNER performance with minimal human effort. Existing research highlights the importance of incorporating candidate entity type information for accurate entity recognition and classification by LLMs. However, when too many candidate entity types are provided in the prompt, LLMs struggle to accurately recognize and label entities in scientific texts, where entity types are more complex than in general domains. To address this challenge, we propose TdSciNER, a **t**ype-**d**riven approach that effectively leverages entity type information to enhance **SciNER** performance. In TdSciNER, we first design an entity type filter model to identify the most likely entity types present in a given sentence. Subsequently, we introduce an auxiliary multi-class entity typing task within a multi-task learning framework alongside SciNER to obtain richer contextual representations. Then, we develop a novel demonstration selection strategy based on sentence similarity and entity type diversity to activate the in-context learning capabilities of LLMs, thereby improving entity recognition accuracy across diverse scientific domains. Experiments on three datasets demonstrate that our method achieves performance comparable to fully supervised models. Further analysis validates that each entity type-driven component in TdSciNER contributes to the improvement of SciNER performance. This work provides valuable insights for future advancements in SciNER and broader information extraction tasks in scientific text mining.

---

* Corresponding author: Chengzhi Zhang (zhangcz@njust.edu.cn).

# 1. Introduction

With the rapid growth of scientific publications, automatically extracting valuable information from large-scale scientific literature has become increasingly important. A crucial step in the scientific information extraction process is named entity recognition, which aims to identify entities such as drugs, chemicals, and proteins in the biomedical domain. In the age of data-driven science, scientific named entity recognition (SciNER) plays an important role in extracting fundamental knowledge units from scientific texts and supporting various downstream applications, such as intelligent answering system (Hu & Ma, 2023; Li et al., 2019), information search (Brandsen et al., 2022), and the development of knowledge graphs (Xie et al., 2016).

SciNER has received widespread attention in fields such as biomedical science (Crichton et al., 2017), computer science (Hou et al., 2019), and materials science (Polak & Morgan, 2024) over the past decade. To enhance SciNER performance, great efforts have been made to optimize methods based on advances in natural language processing (NLP) techniques. Generally, existing SciNER methods can be broadly classified into two main categories: traditional deep learning-based methods and pre-trained language model (PLM)-based methods. Traditional deep learning methods typically employ sequence modeling architectures for representation learning and have achieved promising results. For example, Habibi et al. (2017) introduced a sequence tagging framework that incorporates word embedding layers and recurrent neural networks to recognize biomedical entities. However, deep learning methods often require a large amount of training data, which restricts their performance in low-resource settings. In contrast, PLM-based methods such as BERT (Devlin et al., 2019) are trained on large-scale datasets and allow them to obtain rich contextual representations, which can be effectively fine-tuned for downstream tasks with limited additional training data. For instance, domain-specific models like SciBERT (Beltagy et al., 2019) and BioBERT (Lee et al., 2020), which are pre-trained on scientific and biomedical texts, respectively, have shown superior performance over traditional deep learning methods in domain-specific SciNER tasks. However, these models still require high-quality annotated data to achieve optimal performance in domain-specific settings, which limits their ability to generalize across disciplines.

The development of large language models (LLMs), such as GPT-4, has led to notable progress in various NLP tasks, presenting new opportunities to further optimize entity recognition with minimal human effort. This shift can be attributed to the emergence of in-context learning (ICL) capabilities in latest LLMs, enabling them to understand and execute NLP tasks with only a few provided demonstrations (Brown et al., 2020). Recent studies indicate that LLMs, through few-shot learning, can perform SciNER across multiple disciplines and achieve performance comparable to fully supervised methods (Hu et al., 2023; Peng et al., 2024; Polak & Morgan, 2024). However, it is important to note that SciNER, formulated as a sequence labeling task within named entity recognition (NER), differs from the generative nature of most decoder-based LLMs, which are primarily designed for generative tasks. This mismatch requires tailored prompts to adapt the generative capabilities of LLMs to the sequence labeling demands of SciNER. As shown in Figure1(**a**), in current SciNER research, prompts are typically organized into two components: **(1)** task instructions, which guide the model in understanding and performing the task, and **(2)** output options, which list all candidate entity types and constrain the model's predictions to the predefined label set.

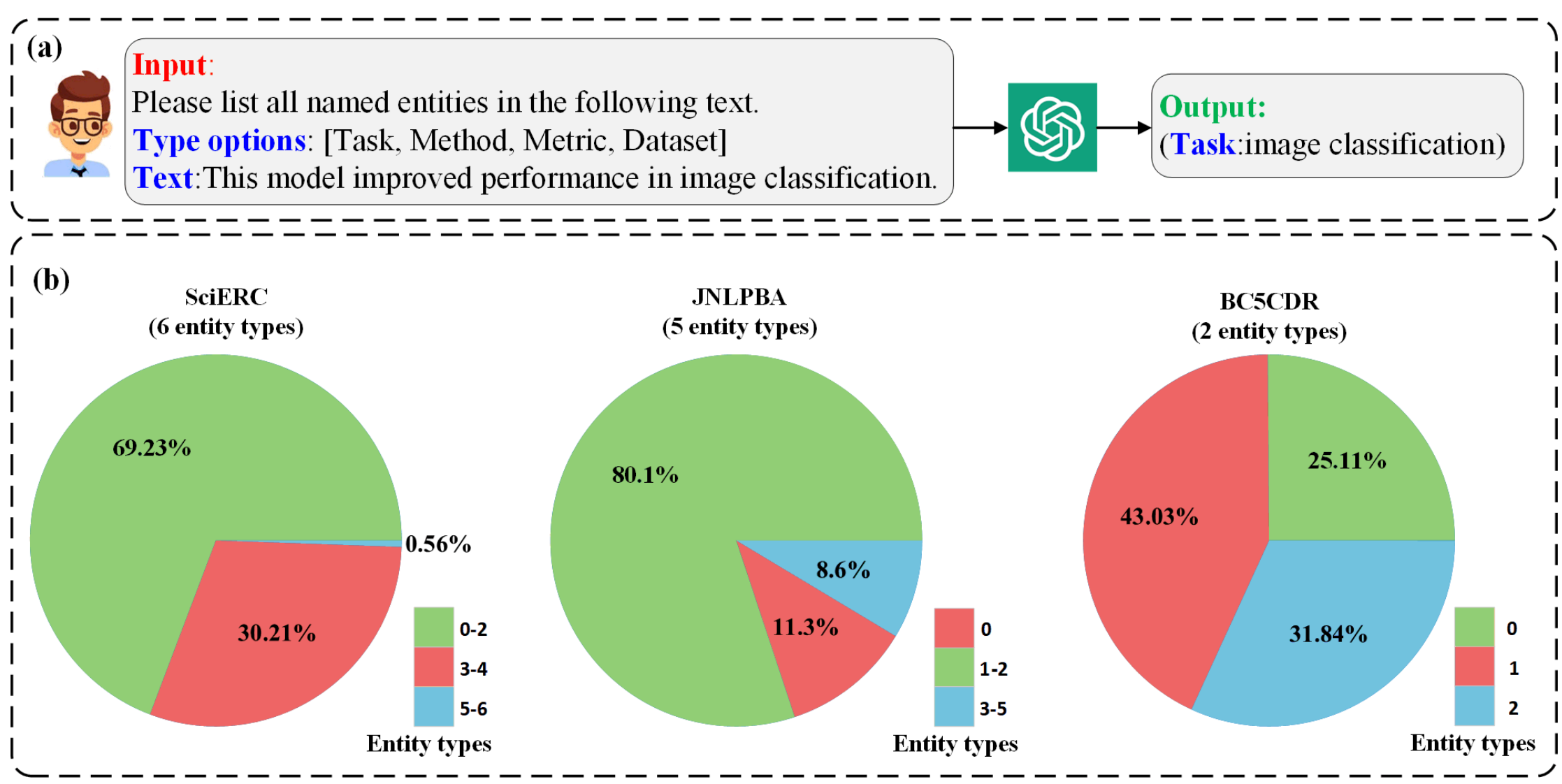

**Figure 1. The general workflow for executing SciNER tasks using LLMs and entity type distribution across SciNER datasets. (a) Standard prompt for SciNER with LLMs. (b) Distribution of entity types per sentence in three SciNER datasets: SciERC (Computer Science), BC5CDR and JNLPBA (Biomedical). In all three datasets, most sentences contain fewer than two entity types, which is significantly lower than the total number of candidate entity types.**

However, as shown in Figure1(**b**), our analysis of three widely used SciNER datasets reveals that most scientific sentences contain only a subset of the entity types present in the full dataset. In such cases, providing irrelevant candidate types in the prompt not only confuses LLMs but also increases the risk of misclassifying scientific entities. In the general domain, several strategies, such as incorporating entity type embeddings, entity type features, or entity type descriptions have been shown to improve NER performance (Feng et al., 2023; Zheng et al., 2021). Despite these successes, such methods are difficult to apply directly to SciNER, where entity recognition is more challenging due to the higher specialization of scientific domains and the need for stronger cross-domain adaptability. For example, in computer science, the term "*Method*" broadly refers to an algorithm or program used to solve a task, such as a sorting algorithm or a machine learning training process (Hou et al., 2019). In contrast, in the biomedical domain, "*Method*" is more narrowly defined as the specific steps or procedures involved in drug trials or experimental protocols (Lee et al., 2020). These differences illustrate that SciNER across disciplines requires substantial entity-related information and domain-specific expertise to accurately distinguish entities. In particular, the absence of domain-specific entity type information poses two challenges for LLMs in completing SciNER tasks: **(1)** difficulty in distinguishing closely related entity types within the same domain, and **(2)** confusion in identifying entities across domains due to limited domain knowledge. Therefore, entity type information holds significant potential for improving LLM-based SciNER. To our knowledge, no previous work has explored how to leverage entity type information in LLMs to enhance SciNER.

To address these gaps, we propose TdSciNER, a type-driven model that leverages entity type information to enhance LLM performance in SciNER. First, we design an entity type filter to estimate the matching degree between input sentences and candidate entity types, which helps identify the entity types that are more likely to be present in a sentence. This process enables the LLMs to concentrate on the relevant entity types during the implementation of the SciNER task. Furthermore, during the fine-tuning stage, we introduce an auxiliary entity typing task and jointly train it with SciNER in a multi-task learning framework to enrich contextual representations and further improve SciNER results. It should be noted that the auxiliary task does not need any further annotations, as it only needs to predict the entity type of a given entity.

Finally, inspired by prior work highlighting the effectiveness of ICL in low-resource information extraction (Hu et al., 2023; Peng et al., 2024; Polak & Morgan, 2024), we develop a similarity-diversity balanced sample selection strategy to select suitable annotated examples as few-shot demonstrations. By employing few-shot learning during inference, we activate the ICL capabilities of LLMs, enabling the model to better grasp the task requirements and achieve more accurate SciNER results. Our contributions can be summarized as follows:

**(1) First**, we propose a new type-driven SciNER model called TdSciNER, which leverages entity type information to guide LLMs for improved SciNER results. To the best of our knowledge, TdSciNER could be the first approach to utilize entity type information within LLM-based SciNER.

**(2) Second**, to implement TdSciNER, we not only develop an entity type filter model to identify the entity types that are most likely to appear in the input sentence, but also design an auxiliary entity typing task to enrich contextual representations through multi-task learning. Both contributions significantly improve the performance of TdSciNER.

**(3) Third**, we introduce a similarity-diversity selection strategy to find labeled samples as demonstrations, which are then used to trigger the ICL capabilities of LLMs and obtain enhanced SciNER results.

**(4)** Finally, experiments conducted on three datasets from different domains have demonstrated that our model achieves results comparable to fully supervised methods. Additionally, each component within TdSciNER is validated to positively contribute to the overall performance of SciNER.

The source code and dataset are publicly available at: https://github.com/tongbao96/code-for-SciNER.

# 2. Related work

In this study, we propose TdSciNER, which leverages entity type information and a multi-task learning framework to improve SciNER performance. In this section, we first review the related work on SciNER. Then, we summarize prior studies on in-context learning with LLMs for advanced NER. Finally, we provide an overview of multi-task learning in information extraction.

## 2.1 Scientific entity recognition

In the general domain, named entities typically include categories such as people, organizations, locations, and dates (Li et al., 2020). In contrast, scientific domains involve a broader and more specialized set of entity types. For example, biomedical entities include genes, proteins, diseases, and drugs, whereas task, model, and dataset are commonly entity types in computer science (Wang et al., 2022c). Accordingly, developing effective methods for SciNER has long been an important research topic. For instance, Swain & Cole (2016) developed an entity recognition tool for chemistry, where entities are labeled with chemical properties, spectral attributes and experimental data. In computer science, Hou et al. (2021), Yan et al. (2020),Yao et al. (2023), Zhang and Ren (2020), and Zhang et al. (2024) extracted tasks, datasets, and evaluation metrics from experimental literature and further predicted relationships between these entities. In the biomedical field, numerous studies have focused on recognizing entities to support the construction of knowledge graphs and question-answering systems (Hu et al., 2024; Liu et al., 2017; Luo et al., 2020; Zhang et al., 2018). To further improve SciNER performance, researchers have also fine-tuned pre-trained models on domain-specific datasets. Representative examples include SciBERT (Beltagy et al., 2019), BioBERT (Lee et al., 2020), and BatteryBERT (Huang & Cole, 2022), which are designed for scientific text, biomedical text, and battery-related literature, respectively.

In recent years, the emergence of LLMs has greatly enhanced the ability to extract information from unstructured scientific papers. Dagdelen et al. (2024) proposed using LLMs to jointly extract named entities and relationships from scientific texts. Similarly, Polak and Morgan (2024) introduced ChatExtract, which utilizes prompt engineering to guide LLMs in automatically extracting entity information from material field. Luiggi et al. (2024) proposed CALM, a context augmentation method which defines prompts as pairs of specific tasks and their corresponding response strategies, demonstrating strong performance in LLM-based NER. Additionally, Peng et al. (2024) employed a soft prompt-based method for extracting clinical concepts and relationships, showing that LLMs with more parameters exhibit greater potential for improved information extraction. In summary, although entity recognition in general domains has achieved impressive accuracy, recognizing entities from scientific papers remains a challenge due to the fine-grained entity types and domain-specific terminology. Although several studies have applied advanced LLMs, such as ChatGPT, to improve SciNER performance, these models still lag supervised or fine-tuned domain-specific models in certain benchmark settings (Wang et al., 2025a; Naguib et al., 2024). Table **1** presents a summary of representative works for SciNER.

**Table 1. Summary of representative studies for SciNER.**

| Papers | Model/ Algorithm | Domains | Main Contributions |
|---|---|---|---|
| Liu et al. (2017) | LSTM+CRF | Clinical | Recognition from clinical text using deep learning methods, achieving over 90% accuracy on multiple i2b2 datasets without the need for feature engineering. |
| Beltagy et al. (2019) | SciBERT | Scientific | A model trained on scientific literature based on BERT, facilitating advancements in various scientific domain NLP tasks. |
| Lee et al. (2020) | BioBERT | Biomedical | BioBERT significantly improves performance on biomedical-related NLP tasks compared to the general BERT model. |
| Huang & Cole (2022) | BatteryBERT | Battery technology | BatteryBERT is pre-trained on a large corpus of battery-related research papers, patents, and technical documents. |
| Huang et al. (2023) | FinBERT | Financial | FinBERT is regarded as the SOTA pre-trained model in the financial domain. |
| Polak & Morgan (2024) | ChatExtract | Materials | This paper introduces ChatExtract, a unified framework for material information extraction based on generative LLMs. |
| Dagdelen et al. (2024) | LLM-NERRE | Scientific | This paper extracts structured named entities and relationships between entities from scientific texts by leveraging advanced LLMs. |

Different from previous studies, our work systematically incorporates entity type information into LLM-based SciNER. Specifically, we develop three entity type-driven components. First, an entity type filter is constructed to identify the entity types most likely to appear in a given sentence. Second, an auxiliary entity typing task is introduced to enhance contextual representation learning for the main SciNER task. Third, both sentence similarity and entity type diversity are considered during demonstration selection to better activate the in-context learning capabilities of LLMs. Experimental results show that these three components jointly contribute to the improvement of SciNER performance.

## 2.2 In-context learning for NER

In-Context Learning (ICL) is an emergent capability of LLMs, enabling them to perform a wide range of downstream tasks using only a few example demonstrations without parameter tuning (Brown et al., 2020). Existing studies have explored various ways to enhance the ICL capabilities of LLMs, including prompt design (Liu et al., 2022; Wang et al., 2023a), demonstration generation (Rubin et al., 2022), and prompt ranking (Zhou et al., 2022). In the context of NER, several studies have attempted to adapt LLMs to entity recognition tasks through ICL. For instance, Lu et al. (2022) proposed UIE, a universal information extraction approach that employs a type-based method to extract entities from diverse domains with higher precision. Wang et al. (2022a) introduced a two-stage framework that integrates span extraction and mention classification to identify entity boundaries and their corresponding types. To better align LLMs with the training objective of NER tasks, Wang et al. (2023b) proposed GPT-NER, which employs a self-assessment method to tackle the problem of overly confident outputs produced by LLMs. Santoso et al. (2024) proposed using open-source LLMs to generate NER data from only a few labeled examples, reducing the reliance on extensive human annotation. Similarly, Jiang et al. (2024b) proposed ToNER, an approach that first identifies candidate entity types within the text, and then leverages ICL to optimize LLMs for general-domain NER. Another research direction is to explore the influence of prompt selection on NER results. Wan et al. (2023) demonstrated that high-quality prompts enriched with reasoning logic can improve ICL performance in relation extraction tasks. Zamai et al. (2024) further found that incorporating definitions and guidelines into prompts helps LLMs better understand task requirements, which is beneficial for improving performance in various information extraction tasks. Table **2** summarizes representative works on in-context learning for NER.

The previous studies provide valuable insights into combining ICL to improve NER performance. However, existing demonstration selection strategies often select demonstrations based on general criteria such as textual similarity, retrieval relevance, or example diversity. These strategies are mostly developed for general-domain NER, where entity types and linguistic expressions are less specialized than those in scientific texts. As a result, they may not fully address the specific challenges of SciNER tasks. Scientific papers often contain specialized jargon, uncommon phrases, and domain-specific expressions that rarely appear in general-domain texts. Moreover, SciNER involves finer-grained and more specialized entity types, making it more challenging than NER in general domains. Therefore, without sufficiently domain-relevant and representative demonstrations, LLMs may struggle to understand specialized scientific language and accurately identify named entities in scientific texts.

To address these issues, we design a dynamic demonstration selection process that jointly considers sentence similarity between demonstrations and the input text, the diversity of represented entity types, and the structural variety of demonstrations. Appropriate weights are dynamically assigned to these criteria to maximize ICL performance. This approach ensures that the selected demonstrations are not only well matched to the input context but also cover a broader range of entity types and structural patterns. Experimental results demonstrate that the examples selected by our strategy exhibit greater robustness than those selected randomly or solely based on similarity or diversity criteria.

**Table 2. Summary of representative works on in-context learning for NER.**

| Papers | Model/ Algorithm | Main Findings |
|---|---|---|
| Wang et al. (2021a) | SpanNER | Combining span-based entity extraction combined with natural language supervision in the prompt helps improve NER performance. |
| Wang et al. (2022a) | SpanProto | This paper highlights the advantages of a span-based approach to achieve improved accuracy in NER in scenarios with limited annotated data. |
| Das et al. (2022) | CONTaiNER | Optimizing the spatial distribution distance of NER entities through contrastive learning techniques can prevent overfitting and improve the model's generalization ability. |
| Xia et al. (2023) | RerankNER | This paper designs a reranking framework that leverages decoded candidates as supervision to debias generative NER. |
| Zamai et al. (2024) | SLIMER | The incorporation of entity definitions and task-specific guidelines into prompts leads to better NER performance, especially in labeling unseen named entities. |
| Jiang et al. (2024a) | P-ICL | Clustering entity types and using point entities as auxiliary information to enhance the performance of ICL in NER tasks. |
| Zhu et al. (2024) | GL-NER | This paper indicates that incorporating specific guidelines into the prompt provides clearer context for classification and decision-making during the few-shot NER process. |

### 2.3 Multi-task learning in information extraction

Multi-task learning aims to train a model on several interrelated tasks simultaneously, thereby improving learning efficiency and enhancing the model's generalization capability. Previous studies have shown that, in information extraction tasks, designing related auxiliary tasks can contribute to improving the outcomes of the primary task (Zaporojets et al., 2021; Hang et al., 2021; Zhang et al., 2018; Kruengkrai et al., 2020; Luo et al., 2024). For instance, Yang and Mitchell (2016) proposed a joint framework that integrates NER and event extraction as joint tasks to achieve cross-document information extraction. Similarly, Martins et al. (2019) introduced entity linking as an auxiliary task alongside the main NER task, focusing on linking recognized entities to a knowledge base to enhance NER performance. In scientific entity recognition, Crichton et al. (2017) incorporated part-of-speech tagging as an auxiliary task to support biomedical entity recognition. Luan et al. (2017) introduced a sequence tagging method for extracting key-phrase from scientific articles, incorporating semi-supervised methods to tackle the challenges posed by limited annotated data. Building on these findings, Luan et al. (2018) further presented a unified approach for extracting relations and coreference clusters from AI conference papers. Weston et al. (2019) developed an information extraction pipeline to extract entities from materials science, and improved recognition performance through entity normalization. Moreover, recent research has examined leveraging LLMs to boost the performance of multi-task learning. Techniques such as instruction tuning (Wang et al., 2023c) and few-shot learning (Wang et al., 2022b; Xie et al., 2023; Lee and Kim, 2023) have been investigated to maximize the capabilities of LLMs in multi-task learning settings.

In summary, multi-task learning has shown considerable potential for improving the effectiveness of information extraction tasks. Existing research studies employ deep learning methods to design auxiliary tasks, which often require additional data annotation. In contrast, advanced LLMs provide a more flexible way to perform multi-task learning for information extraction with only a few prompt examples. However, these efforts have primarily focused on general domains, leaving a research gap in applying multi-task learning to scientific information extraction tasks.

To address this gap, we propose a multi-task learning framework for SciNER that combines the main entity recognition task with an auxiliary multi-class classification task to improve recognition accuracy. In the auxiliary task, we reuse the same annotations as those used for SciNER, thereby avoiding additional labeling costs. This strategy not only simplifies the training process but also enhances the model's ability to learn contextual representations, ultimately improving SciNER performance.

# 3. Methodology

In this paper, we propose TdSciNER, a type-driven SciNER framework that combines an entity type filter with a multi-task learning strategy to optimize LLMs for enhanced SciNER performance. The overall framework of TdSciNER is shown in Figure **2**.

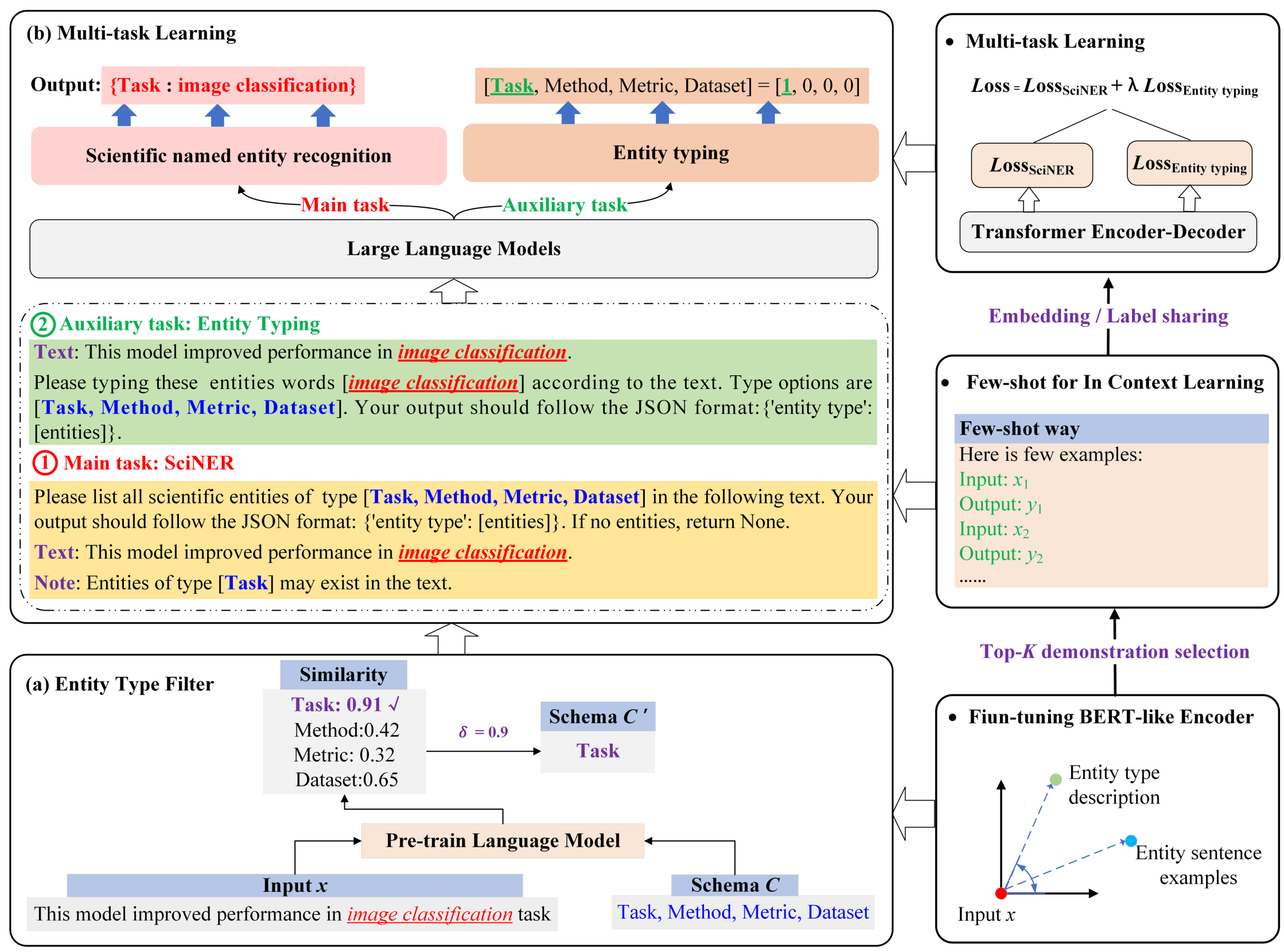

**Figure 2. The overall framework of the proposed TdSciNER.**

The TdSciNER framework includes two main components: **(1)** an entity type filter and **(2)** a multi-task learning module. The entity type filter is designed to compute the matching degree between an input sentence and potential entity types. Then, a threshold is applied to filter out irrelevant entity types and retain those that are more likely to appear in the sentence. As a result, the candidate entity-types pool provided to LLMs is narrowed, which helps the model to concentrate on the most relevant entity types during the SciNER task. In the multi-task learning module, we introduce an entity typing task as an auxiliary task to SciNER, aiming to guide the LLMs in correctly assigning types from the candidate list. Through multi-task learning, TdSciNER leverages shared encodings across tasks and obtains richer contextual understanding without requiring additional data annotation. Finally, we implement a relevance-diversity selection strategy to identify optimal demonstrations that leverage the ICL capabilities of LLMs, helping the model better understand task requirements and achieve enhanced SciNER results. The notation and abbreviations used in our method are listed in Table **3**.

**Table 3. Summary of notation and abbreviations in this section.**

| Notation or Abbreviations | Description |
|---|---|
| $x$ | Input scientific text |
| $I$ | Instruction for LLMs to complete the NER task |
| $T$ | Output of LLMs containing entities and entity types |
| $f_{ETF}$ | Entity filter model |
| Set $C = \{c_1, c_2, \ldots, c_n\}$ | Candidate entity type set |
| Set $D = \{d_{c_1}, d_{c_2}, \ldots, d_{c_n}\}$ | Entity type description set |
| Set $S = \{S_{c_1}, S_{c_2}, \ldots, S_{c_n}\}$ | Sentence set containing a single entity type |
| Threshold $\delta$ | Entity type similarity filtering threshold |
| Set $C'$ | Filtered entity type set |
| Set $K_{candidate} = \{k_1, k_2, \ldots, k_m\}$ | Candidate set for selecting demonstrations |
| $\mathcal{L}$oss | Loss function |
| w, λ, α, $\beta$, $\gamma$ | Control parameters |

### 3.1 Task formalization for SciNER

Scientific named entity recognition (SciNER) is a sub-task of NER that focuses on identifying and classifying entities in scientific text. Formally, given an input scientific text *x*, where *x* is a sequence of tokens, including words or sub-words, we manually design an instruction *I* to guide LLMs in performing SciNER. The instruction typically includes: **(1)** a precise task description, **(2)** a list of entity types to be recognized, and **(3)** the expected format for the model, such as JSON-like entity-type pairs. Then, the instruction *I* is concatenated with the input text *x* to form the full prompt:

$$\text{Prompt } = I \oplus x \tag{1}$$

where $\oplus$ denotes concatenation.

The LLM takes the full prompt as input and generates an output token sequence $y = \{y_1, y_2, \ldots, y_L\}$, where $L$ represents the length of the generated sequence. This output encodes the recognized scientific entities $e_i$ and their corresponding types $c_i$, where $e_i$ denotes the $i$-th entity and $c_i$ its associated type. The output sequence $y$ is structured to represent a set of entity-type pairs:

$$T = \{(c_1, e_1), (c_2, e_2), \dots, (c_m, e_m)\} \quad (2)$$

For example, suppose the input sentence contains the phrase "*Named Entity Recognition*" and the predefined entity types include [*Task*, *Method*, *Metric*], the LLM output should include an entity-type pair such as *("Task", "Named Entity Recognition")*. The generation process is modeled as a conditional probability distribution:

$$p(y \mid x, I) = \prod_{i=1}^{L} p(y_i \mid y_{<i}, x, I) \quad (3)$$

where each token $y_i$ is generated based on all previously generated tokens $y_{<i}$, the input text $x$, and the instruction $I$.

The training objective is to maximize the likelihood of generating the correct output sequence $y$. Accordingly, we minimize the negative log-likelihood loss as follows:

$$\mathcal{L}oss_{\text{SciNER}} = -\sum_{i=1}^{L} \log p(y_i \mid y_{<i}, x, I) \quad (4)$$

During training, model parameters are updated to minimize this loss using ground-truth labeled data with annotated entity spans and types. During inference, the LLM generates the output sequence according to the given prompt, which is then parsed to extract the predicted entities and their corresponding types.

### 3.2 Entity type filter model

In supervised model-based NER, entities are annotated with predefined labels, and the model learns to detect both entity boundaries and their corresponding types. In contrast, generative models generate text based on user-provided prompts. Therefore, a complete list of candidate entity types is usually provided in the prompt to guide the model in assigning type labels to the recognized entities. However, providing too many candidate entity types increases task complexity, which may confuse the model during type assignment and lead to reduced recognition accuracy. To address this, we introduce an entity type filter model denoted by $f_{ETF}$, to reduce the number of candidate entity types presented to LLMs.

**Table 4. Entity descriptions and corresponding examples for the SciERC dataset.**

| Entity Type | Description | Examples |
|---|---|---|
| Task | Refers to specific applications, problems to solve, or systems to construct within a particular domain. | Content summarization, Non-rigid image registration |
| Method | Refers to the various techniques, models, systems, tools, components, or frameworks utilized to achieve a task. | Language modeling, Linear combination |
| Metric | Refers to the standards, measures, or criteria applied to evaluate the quality, performance, or effectiveness of methods and tasks. | Accuracy, Recall, BLEU score |
| Material | Includes data, datasets, resources, corpora, or knowledge bases that are essential for conducting experiments or developing models. | Conversational speech, EUROPARL corpus |
| Generic | General terms or pronouns used as connecting words within texts. These are not domain-specific but serve essential functions in the structure and flow of writing. | algorithm, patterns, framework |
| Other-Scientific-Term | Refers to specific phenomena, concepts, or conditions that influence research outcomes but are not directly related to tasks, methods, metrics, or materials. | Noise, Untextured objects, Mole |

Assuming the original entity-types set is denoted as $C = \{c_1, c_2, \dots, c_n\}$. Since a single entity type name $c_i$ ($1 \leqslant i \leqslant n$) provides limited information for accurately estimating the similarity between the input text and candidate entity types, we use GPT-4 to generate domain-specific descriptions for all entity types. This results in an entity type description set $D = \{d_{c_1}, d_{c_2}, \dots, d_{c_n}\}$, where each description $d_{c_i}$ provides detailed semantic information about the corresponding entity type $c_i$. Notably, the descriptive information in $D$ varies across datasets, domain-specific contexts and semantic nuances of different entity types.

Next, we randomly selected a small subset from the original datasets to construct an entity type example set $S = \{S_{c_1}, S_{c_2}, \dots, S_{c_n}\}$, where each subset $S_{c_i}$ contains annotated texts associated with only one entity type $c_i$. This design allows each subset to capture the semantic characteristics of a specific entity type while avoiding interference from other entity types. Both the entity type description set $D$ and the example set $S$ provide richer semantic information, which helps improve the accuracy of $f_{ETF}$ in estimating the relevance between input sentences and candidate entity types. Taking the SciERC dataset as an example, we provide entity descriptions and corresponding examples in Table **4**.

Subsequently, we employ a BERT-like encoder in $f_{ETF}$ to encode the input sentence. For a given input text $x$, the vector representation is denoted as:

$$h_x = Encoder(x) \tag{5}$$

Then, we first compute the cosine similarity between $h_x$ and each entity type description $d_{c_i}$ in the description set $D$:

$$f_{ETF}\left(h_x, d_{c_i}\right) = \frac{{h_x}^T d_{c_i}}{\|h_x\|_2 \|d_{c_i}\|_2} \tag{6}$$

Similarly, we compute the cosine similarity between $h_x$ and each subset $S_{c_i}$ in the entity type sentence set $S$, denoted as $f_{ETF}\left(h_x, S_{c_i}\right)$. The final similarity score between the input sentence and each candidate entity type is obtained by combining the description-based and example-based similarity scores:

$$f_{ETF} = w \cdot f_{ETF}\left(h_x, d_{c_i}\right) + (1 - w) \cdot f_{ETF}\left(h_x, S_{c_i}\right) \tag{7}$$

where $w$ represents the weights assigned to different similarity scores.

Finally, a threshold $\delta$ is applied to filter out entity types with similarity scores lower than the threshold, resulting in a filtered candidate entity type $C'$. Compared with the original entity type set $C$, $C'$ contains fewer and more relevant entity types for the input sentence.

Based on $C'$, the prompt for LLMs is constructed as shown in Figure 3 in the Appendix. To ensure that no correct entity types are overlooked, we first instruct LLMs to identify all potential scientific entities in the input text. Then, the filtered entity type set $C'$ is used as a guiding note to help LLMs focus on the entity types that are more likely to appear in the text. For clarity, the detailed calculation process of the entity type filter model is presented in Algorithm **1**.

| **Algorithm 1** Procedure for entity type filter | |
|---|---|
| **1.** | **Given:** entity type set $C = \{c_1, c_2, \dots, c_n\}$ ; entity description set $D = \{d_{c_1}, d_{c_2}, \dots, d_{c_n}\}$; entity sentence set $S = \{S_{c_1}, S_{c_2}, \dots, S_{c_n}\}$ containing only one entity type; input text $x$; encoder $\mu(\cdot)$; weight $w$; threshold $\delta$ |
| **2.** | **Initialize:** filtered candidate set C' = [] |
| **3.** | **for each** entity type $c_i \in C$ **do** |
| 4. | encode input text $h_x = \mu(x)$ |
| 5. | encode entity description $h(d_{c_i}) = \mu(d_{c_i})$ |
| 6. | encode entity example set $h(S_{c_i}) = \mu(S_{c_i})$ |
| 7. | compute similarity between $h_x$ and $h(d_{c_i})$, $h_x$ and $h(S_{c_i})$ using Eq. (6) |
| 8. | calculate final similarity score with Eq. (7), incorporating the weight $w$ |
| 9. | **if** $sim_{final} > \delta$ **then** |
| 10. | add $c_i$ to the filtered set C' |
| 11. | **end for** |
| 12. | **Output:** Filtered entity set C′ |

### 3.2.1 Fine-tuning BERT-like encoders for entity type filter

To improve the quality of sentence representations and enhance the accuracy of entity type prediction, we fine-tune BERT-based encoders for the entity type filter module. The goal of this fine-tuning step is to enable the encoder to better capture task-specific semantic similarities between input text and candidate entity types.

Given an input sentence $x$, we aim to determine which types from a predefined type set $C$ are most likely to appear in $x$. Specifically, for each sentence, the types of its annotated entities are treated as positive types, while the types in $C$ that do not appear in the sentence are treated as negative types. The model is then trained to assign higher relevance scores to positive types than to negative ones. We formulate this process as a contrastive learning task, where the model learns to distinguish relevant entity types from irrelevant ones for each input sentence.

Formally, let $P_x$ denote the set of entity types present in $x$, and let $N_x$ denote the set of negative entity types that do not appear in $x$. The entity type filter model $f_{ETF}$ is trained using the following loss function:

$$\mathcal{L}oss_{ETF} = -\sum_{t^+ \in \mathcal{P}_x} \log \frac{e^{f_{ETF}(x,t^+)/\tau}}{\sum_{t \in \mathcal{P}_x \cup \mathcal{N}_x} e^{f_{ETF}(x,t)/\tau}} \tag{8}$$

where $t^+$ represents a positive entity type appearing in the input sentence $x$, $t$ represents any candidate entity type from the original dataset, and $\tau$ is a temperature parameter.

### 3.3 Improving SciNER with an auxiliary entity typing task

Previous studies have demonstrated that introducing auxiliary tasks during the fine-tuning stage can improve NER performance (Wang et al., 2022b). Furthermore, the relevance between the primary task and the auxiliary task is crucial for improving the performance of the primary task (Wang et al., 2021b). Thus, we design an auxiliary entity typing task to complement the SciNER. In the auxiliary task, for each training sample from the original dataset, all annotated entities are listed in the prompt, and the LLM is instructed to assign type labels to these entities from the candidate entity types.

Formally, let $T = \{(c_1, e_1), (c_2, e_2), \dots, (c_m, e_m)\}$, where each $e_i$ is an entity extracted from the input text, and $c_i \in C$ is its corresponding entity type from the set of candidate types $C$. We formulate entity typing as an auxiliary task that predicts the entity type for each annotated entity. Given an input text and its entities, the model is instructed to assign a type label from the candidate type set ($C$) to each entity. The auxiliary entity typing task shares the same model parameters as the primary SciNER task and is jointly optimized during training. Moreover, it does not require any additional annotations since entity type labels are already available in the original SciNER datasets.

For each entity $e_i$, the model produces a probability distribution $\hat{p}_i \in \mathbb{R}^{|C|}$ where $\hat{p}_{i,c}$ denotes the predicted probability that entity $e_i$ belongs to type $c_i \in C$. During training, the correct label $c_i$ is treated as the positive class, while all other candidate types $c \neq c_i$ are considered negative. The model is trained using the standard cross-entropy loss:

$$\mathcal{L}oss_i = -\sum_{c \in C} y_{i,c} \log\left(\hat{p}_{i,c}\right) \tag{9}$$

where $y_{i,c} = 1$ if $c = c_i$, and 0 otherwise. The overall loss for a batch is computed as the sum over individual entity losses.

Thus, the overall loss for an input text $x$ can be computed as the sum of the individual cross-entropy losses for each entity in $T$:

$$\mathcal{L}oss_{Entity\ typing} = \sum_{i=1}^{n} \mathcal{L}oss_i \tag{10}$$

Finally, the total loss for TdSciNER is defined as the weighted sum of the primary SciNER loss and the auxiliary entity typing loss:

$$\mathcal{L}oss = \mathcal{L}oss_{SciNER} + \lambda \mathcal{L}oss_{Entity\ typing} \tag{11}$$

where $\lambda$ is a parameter to control the weight of the auxiliary task relative to the main task.

In this way, the auxiliary entity typing task is closely tied to SciNER, as it encourages the model to capture more informative representations of entity types within the text. This further improves the accuracy and robustness of SciNER. The prompt for the entity typing task is shown in Figure 4 in the Appendix.

**Algorithm 2** Procedure for selecting demonstrations

1. **Given:** input text *x*; the candidate example set $K_{candidate} = \{k_1, k_2, \ldots, k_m\}$; number of demonstrations *k*; entity type filter $f_{ETF}$; control weighting factors α, β, γ.
2. **Initialize:** selected demonstration set top-k = []
3. **for each** candidate example $k_i \in K_{\text{candidate}}$ **do**
4. compute semantic similarity $Sim(k_i)$ between *x* and $k_i$ with Eq. (6)
5. compute and normalize the entity type diversity score $div_{type}(k_i)$ using $f_{ETF}$
6. generate syntactic parse trees for *x* and $k_i$
7. compute structural diversity $div_{structure}(k_i)$ between *x* and $k_i$ with Eq. (12)
8. **calculate** combined $Sim(k_i)$, $div_{type}(k_i)$, $div_{structure}(k_i)$ for $k_i$ with Eq. (13)
9. **end for**
10. **Sort examples in descending order.**
11. **Select top-*k* examples.**
12. **Output:** Set of *k* selected examples.

### 3.4 Enhancing SciNER with in-context learning

Since in-context learning (ICL) was first introduced with GPT-3 (Brown et al., 2020), it has become a widely used strategy for efficiently utilizing LLMs, particularly in low-resource scenarios. Previous studies have shown that ICL can enhance the performance of LLMs in NER tasks, and that demonstration selection plays a crucial role in determining the final performance (Wang et al., 2022b; Wang et al., 2023b; Zamai et al., 2024). In these studies, the similarity and diversity of demonstrations to the input text have been consistently identified as key factors. Inspired by these findings, we propose a dynamic demonstration selection process that jointly considers the similarity, entity type diversity, and structural diversity of candidate demonstrations.

First, we randomly select enough examples with moderate sentence lengths from the source training dataset, since extremely short or long sentences may be less effective as demonstrations. This forms the candidate example set $K_{candidate} = \{k_1, k_2, \dots, k_m\}$, where $m$ denotes the total number of selected candidate examples.

Next, using the entity type filter $f_{ETF}$ constructed in Section 3.2, we compute two metrics for each candidate $k_i \in K_{\text{candidate}}$ with respect to the input text $x$ from the test dataset: the cosine similarity between $x$ and $k_i$, and the normalized number of entity types contained in $k_i$. These metrics produce a cosine similarity score $Sim(k_i, x)$ and an entity type diversity score $div_{type}(k_i, x)$ for each candidate example. In addition to entity type diversity, we also consider the structural diversity of candidate examples to ensure diverse syntactic patterns in the selected demonstrations. Specifically, for each candidate $k_i$, we generate a syntactic parse tree and compute the Tree Edit Distance (TED) between the parse tree of the input $x$ and that of $k_i$:

$$div_{structure}(x, k_i) = \text{TED}(x, k_i) = \frac{\text{Edit Distance}(x, k_i)}{max(|x|, |k_i|)} \tag{12}$$

where Edit Distance $(x, k_i)$ refers to the minimum number of insertion, deletion, and substitution operations required to transform the parse tree of $x$ into the parse tree of $k_i$. $|x|$ and $|k_i|$ denote the sizes of the corresponding parse trees, respectively. The structural diversity score ranges from 0 to 1, where values closer to 0 indicate higher structural similarity, and values closer to 1 indicating greater structural difference.

Finally, we combine the entity type diversity score $div_{type}(k_i, x)$, structural diversity score $div_{structure}(k_i, x)$, and cosine similarity score $Sim(k_i, x)$ to rank the candidate examples. The top-$k$ candidate examples are then selected as ICL demonstrations.

$$ICL_{\text{topk}} = \alpha \cdot \text{Sim}(x, k_i) + \beta \cdot \text{div}_{\text{type}}(x, k_i) + \gamma \cdot \text{div}_{structure}(x, k_i) \tag{13}$$

where $\alpha$, $\beta$, and $\gamma$ are control weighting parameters.

After the demonstration selection, the chosen examples are expected to exhibit both high relevance and sentence-level diversity with respect to the input text, which helps LLMs achieve better SciNER performance under ICL settings. A one-shot prompt for SciNER is illustrated in Figure 5 in the Appendix. Algorithm 2 outlines the overall process of demonstration selection.

# 4. Experiment and results

In this section, we detail the experimental setup, including datasets, evaluation metrics, baseline models, and implementation details. Then, we report the results and parameter analysis.

## 4.1 Dataset and evaluation

### 4.1.1 Dataset

In our experiments, we use three widely recognized scientific datasets from computer science and biomedical domains to evaluate the performance of TdSciNER. The statistics for these datasets are presented in Table **5**.

**Table 5. Dataset statistics used in this paper.**

| Dataset | Domain | Types | | #Train | #Valid | #Test |
|---|---|---|---|---|---|---|
| SciERC | CS | 6 | #sentences | 1861 | 275 | 551 |
| | | | #unique entities | 4078 | 654 | 1364 |
| JNLPBA | Bio | 5 | #sentences | 18469 | 1911 | 1922 |
| | | | #unique entities | 19171 | 2154 | 2543 |
| BC5CDR | PubMed | 2 | #sentences | 5141 | 5202 | 5713 |
| | | | #unique entities | 2730 | 2614 | 2727 |

**Note**: "CS" represents "Computer Science" and "Bio" represents "Biomedical."

**(1)** SciERC (Luan et al., 2018): SciERC is a dataset compiled from 500 scientific abstracts collected from 12 AI conferences and workshops. It contains nested entity annotations and includes six entity types.

**(2)** JNLPBA (Collier et al., 2004): JNLPBA is built from the GENIA corpus and consists of abstracts from biomedical research papers. It is annotated with five entity types.

**(3)** BC5CDR (Li et al., 2016): BC5CDR is an annotated relation extraction dataset consisting of 1,500 PubMed documents. It includes annotations for chemicals and diseases and contains two entity types.

### 4.1.2 Evaluation metric

During evaluation, we report the strict entity-level micro-F1 score, which is computed by aggregating true positives, false positives, and false negatives across all entity types. A prediction is considered correct, i.e., a true positive, only when both the entity span and its corresponding type exactly match the ground truth. This evaluation metric is consistent with prior studies, ensuring a fair comparison with existing methods. The formulas for micro-F1 are defined as follows:

$$P = \frac{TP}{TP + FP} \tag{14}$$

$$R = \frac{TP}{TP + FN} \tag{15}$$

$$F_1 = \frac{2 \times P \times R}{P + R} \tag{16}$$

where *TP* (True Positives) represents entities correctly predicted as positive, *FP* (False Positives) refers to entities incorrectly predicted as positive, and *FN* (False Negatives) denotes entities incorrectly predicted as negative.

## 4.2 Baselines and implementation details

### 4.2.1 Baselines

We evaluate TdSciNER against two groups of baselines: **(a)** supervised pre-trained methods that have achieved leading performance across various SciNER tasks, and **(b)** recent LLM-based approaches for direct comparison with our method. For pre-trained model-based methods, we include a variety of representative models and their variants, some of which are pre-trained on large-scale scientific or biomedical corpora:

**(1)** BiLSTM-CRF (Huang et al., 2015) is a classic sequence tagging model widely applied to NER tasks and frequently employed as a baseline in numerous studies.

**(2)** SciBERT (Beltagy et al., 2019) is pretrained on a corpus of 1.14 million scientific papers, consisting of around 82% biomedical articles and 18% computer science articles.

**(3)** BioBERT (Lee et al., 2020) is pre-trained on large-scale biomedical literature, and has achieved state-of-the-art performance in various biomedical NLP tasks.

**(4)** Att-MT-BLLC (Wang et al., 2021a) is a multi-task learning method that utilizes cross-document information to improve NER performance.

**(5)** SpanNER (Ji et al., 2022) is a NER model that integrates the strengths of both sequence labeling and span-based approaches.

**(6)** AMFF (Yang et al., 2022) captures multilevel features from both the current context and surrounding sentences to enhance NER performance.

**(7)** STM (Hou et al., 2024) is a multi-task learning framework incorporating a language model to address data scarcity in biomedical named entity recognition.

**(8)** DMNER (Bian et al., 2024) is a two-stage framework for biomedical NER that leverages external knowledge and semantic similarity matching for entity classification.

**(9)** OpenBIONER (Cocchieri et al., 2025) focuses on a BERT-based cross-encoder architecture that identifies biomedical entities solely based on their descriptions.

For LLM-based methods, we include both open-source and closed-source models to assess the robustness of our approach:

**(1)** GPT-3.5-turbo[1], a widely used chatbot developed by OpenAI, serves as the foundation for numerous AI-driven applications and products.

**(2)** GPT-4[2], the most advanced language model to date, has attracted significant attention from both industry and academia due to its remarkable capabilities across most NLP tasks.

**(3)** UniNER (Zhou et al., 2024) is a unified NER framework based on LLMs that has demonstrated remarkable performance across diverse domains.

**(4)** P-ICL (Jiang et al., 2024a) introduces point entities as type-specific prompts to enhance NER performance with LLMs, addressing the limitations of standard in-context learning.

**(5)** AST (Li et al., 2024) uses adversarial training to identify key weights and create counterfactual examples, which in turn boosts BioNER performance and generalization.

**(6)** TdSciNER (**Ours**): A Flan-T5-based (Chung et al., 2024) generative framework that leverages entity type descriptions to enhanc SciNER.

### 4.2.2 Implementation details

**(1) Details for Fine-tuning BERT-like encoders for entity type filter**

We fine-tune SciBERT on the SciERC dataset and BioBERT on the JNLPBA and BC5CDR datasets, respectively, as SciBERT is pretrained on publications related to computer science and is therefore well aligned with the domain of SciERC, while BioBERT, trained on biomedical corpus, aligns more closely with the latter two biomedical datasets. During fine-tuning, for each training sample, the labeled entity types are treated as positive samples, with all unlabeled types regarded as negative examples. Each model undergoes fine-tuning for 3 epochs with a learning rate of 1e-5, and a batch size of 16. Then, based on the fine-tuning model, we compute the similarity score between the input text and each candidate type (based on its description and example sentences). For the computation, the weights for the entity descriptions and sentences containing only the specific entity type were selected from {0.2, 0.3, 0.4, 0.5} and {0.8, 0.7, 0.6, 0.5}, respectively. The optimal weights were determined to be 0.3 for the entity descriptions and 0.7 for the sentences, as this configuration yielded the best results on the validation set of the subsequent SciNER task.

[1] https://platform.openai.com/docs/models/gpt-3-5-turbo
[2] https://platform.openai.com/docs/models/gpt-4-turbo-and-gpt-4

**(2) Details for Fine-tuning LLMs for SciNER**

For the SciNER implementation, we used Flan-T5-xxl (Chung et al., 2024) as our LLM backbone and fine-tuned it using Low-Rank Adaptation (LoRA) (Hu et al., 2021) to reduce memory consumption by freezing the initial parameters. The fine-tuning was performed on L40 GPUs with 48GB of memory. In preliminary experiments, we tested learning rates of {1e-3,1e-4, 2e-4, 1e-5, 2e-5, 1e-6} and obtained the best performance with 2e-5. The optimal batch size was determined from the options {4, 8, 16, 32}, with a chosen value of 16. The LoRA dropout was set to 0.05, with LoRA-r set to 8 and LoRA-alpha set to 16. The weight for the auxiliary task was selected from {0.04, 0.07, 0.1, 0.2, 0.3}, and 0.1 was found to yield the best trade-off between model performance and training efficiency. Besides Flan-T5-xxl, we additionally fine-tuned two open-source LLMs: Mistral-3B and LLaMA2-7B, to further examine the adaptability of our method across different sizes and architectures. We adopted similar fine-tuning hyperparameters, with minor adjustments (e.g., batch size or learning rate) based on model size. The pre-trained weights for SciBERT, BioBERT and Flan-T5-xxl were sourced from Hugging Face[3], and all models used are the uncased versions. For GPT-3.5-turbo and GPT-4, we evaluated both models using the official API[4] with all parameters set to default values. We adopted the same prompts (see the Appendix for details) as the Flan-T5 series models, along with the demonstration selection process described in Section 3.4. The optimal number of shots was selected from {1, 3, 5, 10, 20}. Due to cost limitations, both models were evaluated in two rounds, and we reported the average results. For demonstration selection during inference, 100

[3] https://huggingface.co/

[4] https://openai.com/api/

candidate samples were taken from the training set to avoid test-time leakage.

### 4.3 Results analysis

In this section, we first present the overall performance comparison across different models. Then, we report the results under different model configurations and parameter settings, followed by insights from the ablation study.

#### 4.3.1 Overall performance comparison

For all baseline models, we report the results on different datasets as presented in their original papers. As shown in Table 6, TdSciNER achieved the highest F1 scores on the SciERC and JNLPBA datasets, while also delivering overall competitive performance compared with strong baselines such as SciBERT and BioBERT. These models are pre-trained on domain-specific corpora and are widely regarded as representative SOTA baselines for SciNER. Since SciERC and JNLPBA correspond to the computer science and biomedicine domains, respectively, this result demonstrates the strong cross-domain adaptability of TdSciNER. On the BC5CDR dataset, although it contains only two entity types, our model also delivered competitive results with an F1 score of 89.83%, which is close to those of strong supervised baselines. This suggests that our model not only performs well in fine-grained entity recognition scenarios but also remains robust when entity-type diversity is limited. In contrast, although commercial LLMs like GPT-3.5-turbo and GPT-4 perform well in general text generation tasks, their performance on SciNER remains limited. Our entity type filtering module brings notable improvements to these models; however, a clear performance gap remains compared with fine-tuned LLM-based approaches. Compared with other open-source fine-tuned methods, TdSciNER significantly outperforms UniNER and P-ICL. Notably, our smaller 3B version even surpasses the larger 7B UniNER model on SciERC and JNLPBA, highlighting the efficiency and effectiveness of our design. Finally, across all three datasets, we observe that TdSciNER achieves greater improvements on datasets with more entity types. This advantage can be attributed to the effectiveness of the entity type filtering module, which eliminates irrelevant types and narrows the candidate type space, thereby enabling the LLM to focus more precisely on relevant entity-type distinctions.

**Table 6. Overall comparison of TdSciNER with baselines on three datasets, where SciERC, JNLPBA, and BC5CDR contain 6, 5, and 2 entity types, respectively.**

| Model | SciERC | | | JNLPBA | | | BC5CDR | | |
|---|---|---|---|---|---|---|---|---|---|
| | P | R | $F_1$ | P | R | $F_1$ | P | R | $F_1$ |
| BiLSTM-CRF （2015） | 67.83 | 47.83 | 56.10 | 73.47 | 68.27 | 70.77 | 88.87 | 86.35 | 87.59 |
| Att-MT-BLLC（2021） | 62.37 | 52.29 | 56.88 | - | - | - | 87.31 | 81.70 | 84.41 |
| **Pre-trained-based methods** | | | | | | | | | |
| SciBERT▲ （2019） | - | - | 67.57 | - | - | 77.28 | - | - | <u>90.01</u> |
| BioBERT▲ （2020） | - | - | - | 72.24 | **83.56** | 77.49 | **92.52** | **92.76** | **92.64** |
| SpanNER （2022） | <u>70.02</u> | **70.91** | <u>70.46</u> | - | - | - | <u>89.37</u> | 90.46 | 89.91 |
| AMFF （2024） | 67.87 | 61.03 | 64.27 | **78.60** | 79.79 | <u>79.19</u> | - | - | - |
| DMNER （2024） | - | - | - | 72.01 | 82.24 | 76.79 | 87.96 | 84.20 | 86.04 |
| STM （2024） | - | - | - | 71.13 | 77.37 | 74.12 | 89.35 | 88.78 | 89.07 |
| OpenBIONER （2025） | - | - | - | - | - | 74.30 | - | - | 86.30 |
| **LLM-based methods** | | | | | | | | | |
| GPT-3.5-turbo (2023) | 22.43 | 19.38 | 20.80 | 27.51 | 49.30 | 35.31 | 45.59 | 54.72 | 49.74 |
| +ETF+ICL | 26.52 | 23.95 | 25.17 | 30.67 | 56.65 | 39.79 | 47.71 | 55.38 | 51.27 |
| GPT-4 (2024) | 30.75 | 22.19 | 25.78 | 31.72 | 51.03 | 39.19 | 45.90 | 56.13 | 50.50 |
| +ETF+ICL | 35.17 | 27.63 | 30.96 | 36.83 | 59.11 | 45.38 | 50.43 | 59.57 | 54.62 |
| P-ICL (7B) (2024) | - | - | 32.17 | - | - | 35.05 | - | - | 62.14 |
| UniNER (7B)‡ (2024) | - | - | 66.00 | - | - | 75.60 | - | - | 89.34 |
| AST (7B) ‡ (2024) | - | - | - | - | - | 79.16 | - | - | 73.93 |
| **TdSciNER (Ours)‡** | | | | | | | | | |
| +Mistral-3B | 67.89 | 66.70 | 67.29 | 73.54 | 78.49 | 76.40 | 87.78 | 89.14 | 88.45 |
| +LLaMA2-7B | 68.92 | 67.57 | 68.24 | 74.95 | 79.82 | 77.31 | 88.45 | 90.29 | 89.36 |
| +Flan-T5-11B | **70.28*** | <u>70.86*</u> | **70.58*** | <u>76.43*</u> | <u>82.48*</u> | **79.34** | 88.96* | <u>90.71*</u> | 89.83* |

**Note**: The best results are in **bold** and the second-best are <u>underlined</u>. The symbol "-" indicates that the respective values are not reported in the referenced paper. "▲" indicates models that are pre-trained on domain-specific datasets and are widely regarded as SOTA. "ETF+ICL" refers to a pure LLM augmented with our two modules: the entity type filter model (ETF) and in-context learning (ICL) using the optimal demonstrations selected by our method. "*" indicates results that show a statistically significant difference compared to the LLM-based baselines. "‡" indicates models that have been fine-tuned on the corresponding dataset.

#### 4.3.2 Encoder analysis for the entity type filter model

In SciNER, textual encoders play a crucial role in transforming input texts into vector representations, which directly affects the overall performance of the model. To examine the influence of different textual encoders, we select seven representative embedding models with different architectures and parameter sizes, including two widely used BERT-like models (BERT-base and BERT-large), two domain-specific models (SciBERT and BioBERT), and three recent sentence embedding models (Sentence-T5, GTE-large, and BGE-large). For the entity type filter model, each encoder is fine-tuned on the corresponding dataset, and its performance is evaluated on the SciNER task under the same generative model setting.

**Table 7. Comparison of encoders for TdSciNER across different datasets ($F_1$ %).**

| Encoder | Domain | #Size | SciERC | JNLPBA | BC5CDR |
|---|---|---|---|---|---|
| BERT-base | General | 110M | 67.92 | 77.45 | 87.68 |
| SciBERT | Scientific | 110M | **70.58** | <u>78.87</u> | <u>89.51</u> |
| BioBERT | Biomedical | 110M | 70.01 | **79.34** | **89.83** |
| Sentence-T5 | General | 168M | 68.24 | 77.94 | 88.24 |
| BERT-large | General | 340M | 69.09 | 78.02 | 88.79 |
| GTE-large | General | 434M | 70.24 | 78.49 | 89.12 |
| BGE-large | General | 438M | <u>70.37</u> | 78.73 | 89.27 |

**Note**: The best results are in **bold** and the second-best are <u>underlined</u>.

As shown in Table 7, domain-specific models such as SciBERT and BioBERT outperform general-purpose models across the SciERC, JNLPBA, and BC5CDR datasets. Specifically, SciBERT achieves the highest F1 score on SciERC (70.58%), while BioBERT achieves the best performance on the biomedical datasets, with F1 scores of 79.34% on JNLPBA and 89.83% on BC5CDR. These results suggest that domain-specific pre-training, particularly on corpora closely related to the downstream task, is beneficial for SciNER. This is because such pre-training enables the model to capture specialized terminology, syntactic patterns, and discourse structures that frequently appear in scientific texts, thereby generating richer contextualized embeddings and achieving better performance than general-purpose models, even those with larger parameter sizes.

#### 4.3.3 Threshold analysis for the entity type filter model

We also investigate the impact of threshold settings across different datasets. In general, a higher threshold reduces the number of candidate entity types but may miss some true entity types. In contrast, a lower threshold improves recall but increases the risk of introducing irrelevant entity types and causing incorrect assignments. Therefore, in a preliminary experiment, we use the entity type filter model $f_{ETF}$ to compute the distribution of positive and negative samples in the training sets of the three datasets. As shown in the Figure 6(**b**), on the JNLPBA dataset, we observed a clear similarity gap between positive and negative samples, with $f_{ETF}$ effectively separating the two sets. On the SciERC (Figure 6(**a**)) and BC5CDR (Figure 6(**c**)) datasets, the similarity distributions overlap more, yet a separation trend is still visible for part of the samples. This is reflected in the probability density plots, where positive pairs tend to receive higher similarity scores, while negative pairs cluster around lower scores.

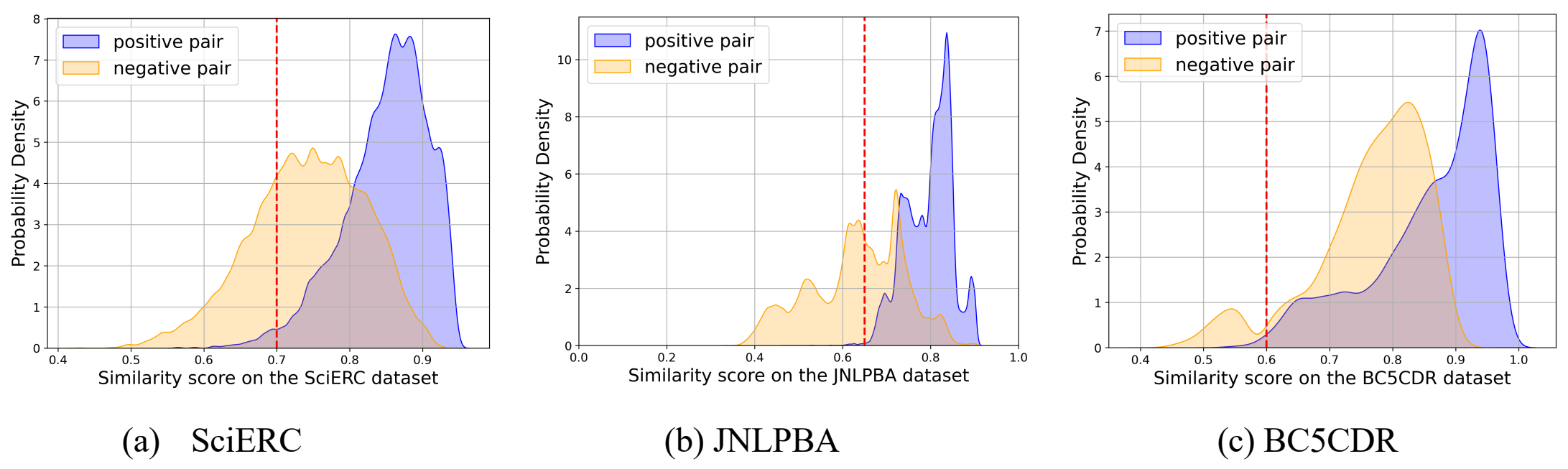

(a) SciERC (b) JNLPBA (c) BC5CDR

**Figure 6. Distribution of positive and negative samples in the training sets across datasets, with the optimal threshold marked by the red line.**

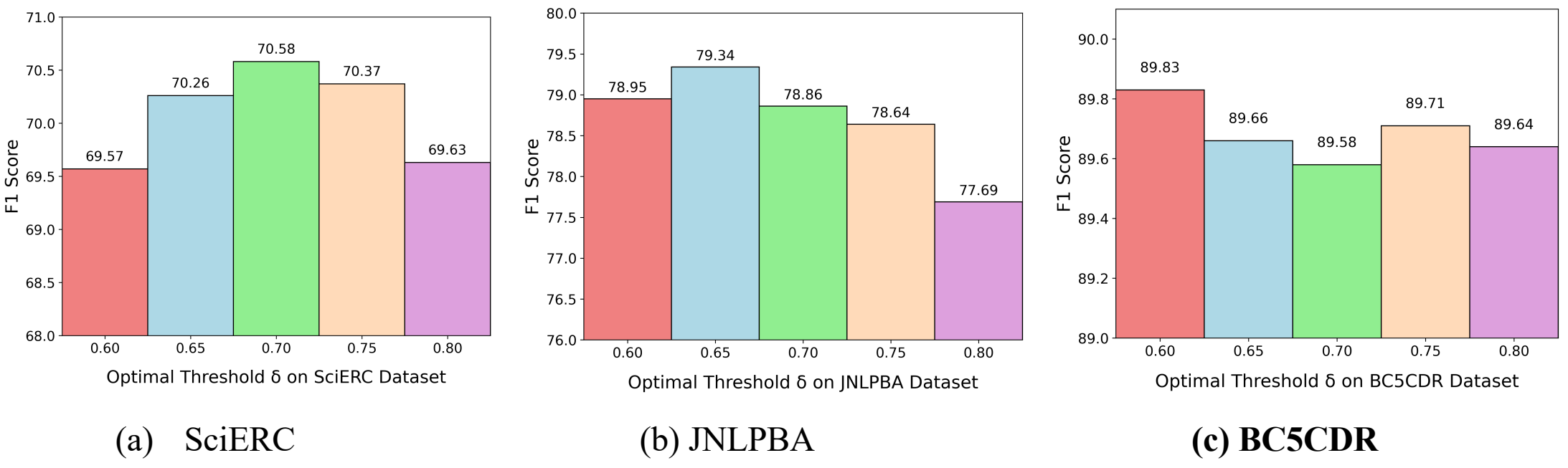

(a) SciERC (b) JNLPBA **(c) BC5CDR**

**Figure 7. Impact of different filtering thresholds across datasets.**

Based on this distribution, we further analyzed the performance of TdSciNER under different threshold values δ. As shown in Figure 7, the optimal δ values for the SciERC, JNLPBA, and BC5CDR are 0.70, 0.65, and 0.60, respectively, indicating that the choice of δ is influenced by the entity type distributions of each dataset. For SciERC, which contains diverse and complex entity types, such as tasks, methods, materials, and metrics, sentences often exhibit semantic overlap among candidate types, especially due to nested entity annotations. Therefore, a relatively higher threshold of 0.70 is needed to more effectively filter out irrelevant candidate types. In contrast, JNLPBA shows a clearer similarity gap between positive and negative samples, with positive samples mainly concentrated between 0.65 and 0.90, while negative samples are distributed between 0.40 and 0.80. Accordingly, a moderate threshold of 0.65 achieves the best balance between retaining relevant types and excluding irrelevant ones. For BC5CDR, which contains only two entity types, the type space is relatively simple, and excessive filtering may remove potentially correct entity types. Therefore, a lower threshold of 0.60 is more suitable.

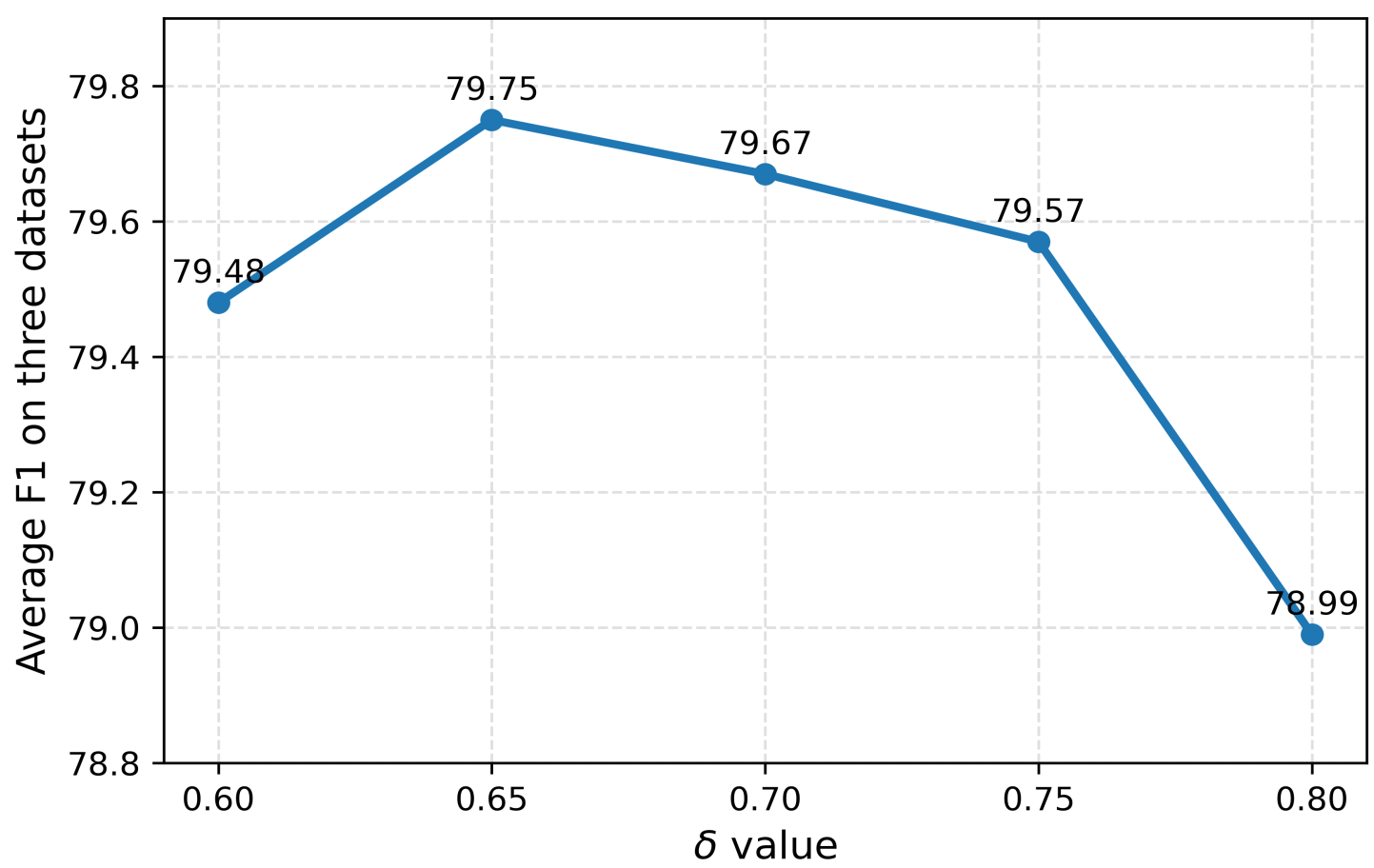

**Figure 8. Cross-domain performance under fixed threshold values**

We evaluate TdSciNER across three datasets using different fixed values of δ. As shown in the Figure **8**, δ = 0.65 achieves the best overall result, with the highest average F1 score of 79.75. Therefore, for applications to unseen domains where no validation set is available, δ ≈ 0.65 can be used as a reasonable default, as it offers a good balance between filtering irrelevant entity types and preserving relevant ones. In addition, for domains involving a larger number of entity types or stronger cross-disciplinary characteristics, a slightly higher threshold may be considered to improve the discrimination among candidate entity types, whereas a lower threshold may be preferable when entity types are limited or highly overlapping.

#### 4.3.4 Demonstration selection strategy analysis for ICL

To explore the impact of different demonstration selection methods on improving TdSciNER's performance, we compare four strategies across the three datasets: random selection, similarity-based selection, diversity-based selection, and a combined similarity-diversity selection strategy. As shown in Table **8**, most similarity-based and combined similarity-diversity settings outperform the zero-shot baseline, indicating that incorporating demonstrations into the prompt can effectively activate the ICL capabilities of LLMs and lead to more accurate SciNER results. Specifically, similarity-based selection achieves better performance than diversity-based selection alone, suggesting that highly similar demonstrations provide more relevant contextual information and help the model better align with the target input. However, when both similarity and diversity are considered, the combined strategy achieves the best overall performance. One possible explanation is that diverse demonstrations introduce a wider range of contexts, entity types, and sentence structures into the prompt, thereby improving the robustness of the model across different scenarios. The results also show that datasets with more entity types, such as SciERC and JNLPBA, benefit more from increasing the number of shots, leading to consistent performance improvements. In contrast, on BC5CDR, which contains only two entity types, the best performance is achieved with 10 shots, and further increasing the number of demonstrations does not bring additional gains.

**Table 8. Effect of demonstration selection strategies on TdSciNER performance ($F_1$ %).**

| Method | #Shots | SciERC | JNLPBA | BC5CDR | Average |
|---|---|---|---|---|---|
| | Zero-shot | 67.41 | 74.85 | 87.96 | 76.74 |
| Random | 1-shot | 67.62 | 75.21 | 87.94 | 76.92 |
| | 3-shot | 68.41 | 75.88 | 88.69 | 77.66 |
| | 5-shot | <u>68.96</u> | 76.91 | 88.95 | 78.27 |
| | 10-shot | 68.74 | <u>77.40</u> | <u>**89.12**</u> | <u>78.42</u> |
| | 20-shot | **69.92** | **77.79** | <u>89.01</u> | **78.91** |
| Similarity | 1-shot | 67.68 | 75.41 | 88.23 | 77.10 |
| | 3-shot | 68.53 | 76.42 | 88.57 | 77.84 |
| | 5-shot | 69.19 | 78.05 | 88.76 | 78.67 |
| | 10-shot | <u>69.97</u> | <u>78.69</u> | **89.58** | <u>79.41</u> |
| | 20-shot | **70.16** | **78.83** | <u>89.42</u> | **79.47** |
| Diversity | 1-shot | 66.36 | 73.07 | 87.01 | 75.48 |
| | 3-shot | 66.97 | 73.62 | 87.43 | 76.01 |
| | 5-shot | 67.34 | 75.34 | <u>87.67</u> | 76.78 |
| | 10-shot | <u>67.51</u> | <u>76.05</u> | **87.91** | <u>77.16</u> |
| | 20-shot | **68.24** | **76.79** | 87.63 | **77.55** |
| Similarity+ Diversity | 1-shot | 67.72 | 75.38 | 88.06 | 77.05 |
| | 3-shot | 68.66 | 76.97 | 88.54 | 78.06 |
| | 5-shot | 69.41 | 78.70 | 89.27 | 79.13 |
| | 10-shot | <u>70.09</u> | <u>79.15</u> | **89.83†** | <u>79.69</u> |
| | 20-shot | **70.58†** | **79.34†** | <u>89.71</u> | **79.87†** |

**Note**: The best results are in **bold** and the second-best are <u>underlined</u>, and the symbol † indicates the best value across all groups.

We further investigate the impact of the demonstration selection ratio on SciNER performance. Taking JNLPBA, the dataset with the largest number of entity types, as an example, we report the performance under different ratio combinations using 20 shots, which is identified as the optimal number of demonstrations for this dataset. As shown in Table **9**, when only two factors are combined, sentence similarity and entity type diversity play more important roles than sentence structure diversity, with higher proportions of the former two generally leading to better entity recognition performance. However, as the weight of sentence structure diversity increases, the performance first improves and then declines. This may be because moderate structural diversity helps the model adapt to diverse linguistic contexts, whereas excessive structural diversity introduces additional noise, making it more difficult for the model to focus on relevant patterns. Overall, the best F1 score of 79.34% is achieved with a ratio of 0.4 for sentence similarity, 0.4 for entity type diversity, and 0.2 for sentence structure diversity. This highlights the importance of balancing multiple factors in the demonstration selection process. For instance, while incorporating sentence similarity and entity type diversity, adding sentence structure diversity ensures that the model not only recognizes entities in familiar sentence forms but also adapts to various structures, thus enhancing the model's robustness and improving SciNER performance.

**Table 9. Effect of demonstration selection ratios on the JNLPBA dataset.**

| **Demonstration Selection Ratio** | | | **P (%)** | **R (%)** | **$F_1$ (%)** |
|---|---|---|---|---|---|
| $\alpha$ | $\beta$ | $\gamma$ | | | |
| 0.5 | 0.5 | - | 76.22 | 82.15 | 79.07 |
| 0.5 | - | 0.5 | 75.27 | 81.04 | 78.05 |
| - | 0.5 | 0.5 | 73.85 | 79.97 | 76.79 |
| 0.2 | 0.3 | 0.5 | 74.92 | 80.09 | 77.42 |
| 0.2 | 0.5 | 0.3 | 75.43 | 81.61 | 78.40 |
| 0.3 | 0.2 | 0.5 | 75.14 | 80.34 | 77.65 |
| 0.3 | 0.5 | 0.2 | <u>76.40</u> | 82.39 | <u>79.28</u> |
| 0.4 | 0.4 | 0.2 | **76.43** | **82.48** | **79.34** |
| 0.4 | 0.2 | 0.4 | 76.03 | 82.15 | 78.97 |
| 0.5 | 0.3 | 0.2 | 76.29 | <u>82.46</u> | 79.26 |
| 0.5 | 0.2 | 0.3 | 76.17 | 82.43 | 79.18 |

**Note**: The best results are in **bold** and the second-best are <u>underlined</u>. '$\alpha$' denotes the sentence similarity between the demonstration and the target sentence. '$\beta$' represents the diversity of entity types in the demonstration, and '$\gamma$' indicates the diversity of sentence structures between the demonstration and the target sentence.

### 4.3.5 Weight selection analysis for multi-task learning

In multi-task learning, the loss weight assigned to the auxiliary task plays a critical role in shaping both training dynamics and the final performance. A weight that is too small may weaken the contribution of the auxiliary task, whereas an excessively large weight may bias the optimization process toward the auxiliary objective, potentially leading to overfitting and degraded performance on the primary task.

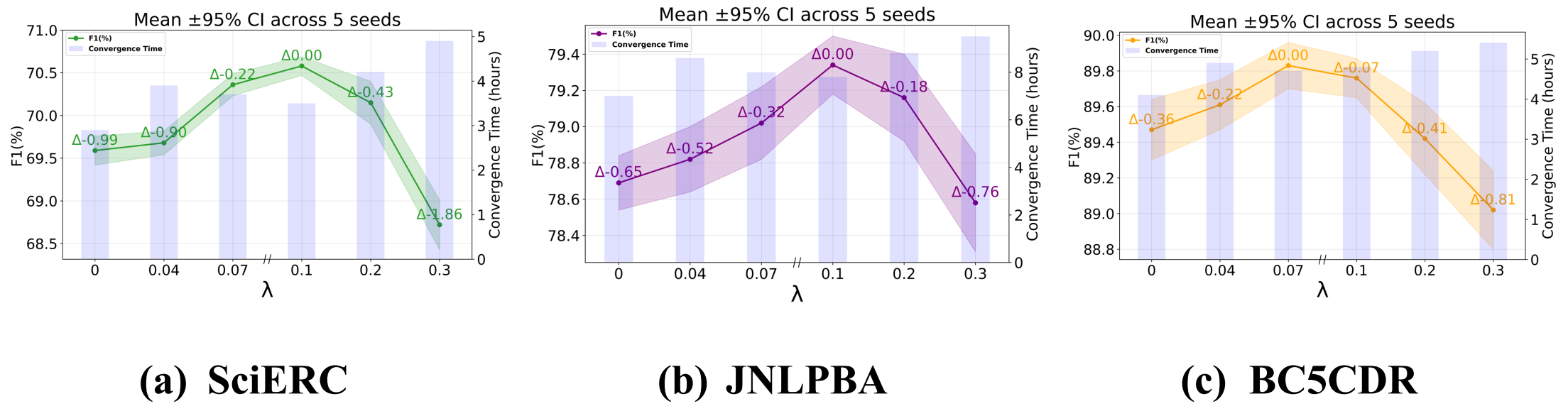

**(a) SciERC** **(b) JNLPBA** **(c) BC5CDR**

**Figure 9. Impact of auxiliary task weights ( λ ) on SciNER performance across datasets.**

As shown in Figure **9**, when the auxiliary task weight is very small, the performance gain is marginal. In contrast, when the weight exceeds a certain threshold, the training time increases and the performance across all datasets **generally** decreases. The optimal weights are 0.10 for SciERC and JNLPBA, which achieve a good trade-off between accuracy and training efficiency, and 0.07 for BC5CDR, where the performance difference between 0.07 and 0.10 is marginal. These results suggest that, for datasets with a larger number of entity types, the auxiliary entity typing task is more beneficial and can be assigned a relatively higher weight. This is because it helps the model better distinguish between closely related entity types and reduces confusion during type assignment. For BC5CDR, which contains only two entity types, the auxiliary task still brings a modest performance gain, but the improvement is relatively limited. Overall, given that the proposed entity typing task shares the same annotations as the primary SciNER task and requires no additional labeled data, these findings further demonstrate the effectiveness and practicality of the auxiliary task design.

### 4.3.6 Computational complexity analysis

Table **10** compares the computational complexity and F1 scores of different LLM-based and LLM-enhanced NER methods. Following prior work, we report the complexity from three stages: sampling, training, and inference. Compared with prompt-based or in-context learning

methods such as GPT-3.5-turbo, GPT-4, and P-ICL, TdSciNER introduces an additional training stage through the proposed multi-task learning framework. Although this increases the training complexity, it brings substantial F1 improvements across all three datasets. For example, compared with P-ICL, TdSciNER improves the F1 score from 32.17% to 70.58% on SciERC, from 35.05% to 79.34% on JNLPBA, and from 62.14% to 89.83% on BC5CDR. Compared with AST, which mainly improves BioNER through adversarial selective training, TdSciNER introduces additional inference-stage operations such as entity type filtering.

**Table 10. Computational complexity and F1 score comparison of LLM-based methods**

| Method | Computational complexity | | | F1 score (%) | | |
|---|---|---|---|---|---|---|
| | Sampling | Training | Inference | SciERC | JNLPBA | BC5CDR |
| GPT-3.5-turbo | - | - | $O(NL)$ | 25.17 | 39.79 | 51.27 |
| GPT-4 | - | - | $O(NL)$ | 30.96 | 45.38 | 54.62 |
| P-ICL (2024) | $O(NL)$ | - | $O(NL)$ | 32.17 | 35.05 | 62.14 |
| AST (2024) | - | $O(NL)$ | $O(NL)$ | - | <u>79.16</u> | 73.93 |
| UniNER (2024) | $O(NL)$ | $O(NL)$ | $O(NL)$ | <u>66.00</u> | 75.60 | <u>89.34</u> |
| TdSciNER (Ours) | $O(NL)$ | $O(NL)$ | $O(NL)$ | **70.58** | **79.34** | **89.83** |

* $N$ denotes the number of processed instances, and $L$ denotes the average input length. For GPT-3.5-turbo and GPT-4, we incorporate ETF and ICL following our proposed method for fair comparison.

To further evaluate the practical efficiency of TdSciNER, we compare the average inference time and recognition performance with representative LLM-based baselines. As shown in Table 11, TdSciNER achieves competitive inference efficiency while obtaining superior NER performance across all three datasets. Compared with GPT-based methods, TdSciNER significantly improves F1 scores while maintaining a comparable inference time. For example, TdSciNER achieves improvements of 49.78%, 44.03%, and 40.09% F1 points over GPT-3.5-turbo on SciERC, JNLPBA, and BC5CDR, respectively, with an inference time of only 2.85 s/sample. Compared with UniNER, TdSciNER obtains consistent performance gains on all datasets, especially on SciERC and JNLPBA, while introducing only a negligible increase in inference time (0.12 s/sample compared to GPT-3.5-turbo). These results demonstrate that the proposed type-driven framework effectively improves SciNER performance without introducing significant inference overhead.

**Table 11. Inference time comparison of TdSciNER and LLM-based baselines.**

| Method | Avg. Inference Time (s/sample) | SciERC (F1%) | JNLPBA (F1%) | BC5CDR (F1%) |
|---|---|---|---|---|
| GPT-3.5-turbo | **2.68** | 20.80 | 35.31 | 49.74 |
| GPT-4 | 3.43 | 25.78 | 39.19 | 50.50 |
| P-ICL (2024) | 5.09 | 32.17 | 35.05 | 62.14 |
| AST (2024) | - | - | <u>79.16</u> | 73.93 |
| UniNER (2024) | <u>2.73</u> | <u>66.00</u> | 75.60 | <u>89.34</u> |
| **TdSciNER (Ours)** | 2.85 | **70.58** | **79.34** | **89.83** |

*The inference time of AST is not reported because the official implementation is unavailable.

Furthermore, we analyze the additional inference overhead introduced by individual components of TdSciNER. As shown in Table **12**, ETF requires 1.02, 1.25, and 1.17 seconds per sample on SciERC, JNLPBA, and BC5CDR, respectively, while bringing consistent F1 improvements. ICL (Sim+Type) introduces a much smaller overhead, requiring only 0.29~0.33 seconds per sample, and still achieves clear performance gains. When further incorporating TED-based structural diversity, ICL (Sim+Type+TED) achieves the largest F1 improvements across the three datasets, although it also incurs higher runtime due to CPU-based TED computation.

**Table 12. Additional inference overhead and F1 improvements of TdSciNER components**

| Method | Inference Time (s/sample) | | | F1 improvements (points) | | |
|---|---|---|---|---|---|---|
| | SciERC | JNLPBA | BC5CDR | SciERC | JNLPBA | BC5CDR |
| ETF | 1.02 | 1.25 | 1.17 | 2.26 | 2.81 | 0.41 |
| ICL (Sim+Type) | 0.31 | 0.33 | 0.29 | 2.21 | 2.75 | 1.61 |
| ICL (Sim+Type+TED) | 1.39 | 1.68 | 1.42 | 2.77 | 3.02 | 1.85 |

**'ETF'* denotes entity type filter model. *'Sim'* denotes cosine similarity between the input sentence and candidate examples. *'Type'* denotes the diversity of entity types across the candidate sentence. The *'ETF'*, *'Sim',* and *'Type'* scores are computed on a local NVIDIA RTX 3090 GPU, whereas TED-based structural diversity score is computed on a local Intel i7-12700K CPU.

### 4.3.7 Ablation study

We investigate the impact of different Flan-T5 backbone sizes on the performance of TdSciNER. In this experiment, we first select the best-performing text encoder from Table **7** as the foundation of the entity type filter model. We then evaluate Flan-T5-small, Flan-T5-base, Flan-T5-large, and Flan-T5-XXL as backbone models for comparison. Since ICL capabilities typically emerge in large-scale models with billions of parameters, the ICL module is applied only to the Flan-T5-XXL model. In addition, we tune the hyperparameters of each model to ensure optimal performance.

**Table 13. Performance comparison of Flan-T5 series models on SciNER tasks ($F_1$ %).**

| Model | #Size | Modules | | | SciERC | JNLPBA | BC5CDR | Average |
|---|---|---|---|---|---|---|---|---|
| | | ETF | AT | ICL | | | | |
| Flan-T5-small | 80M | √ | √ | - | 55.24 | 69.68 | 81.57 | 68.83 |
| Flan-T5-base | 250M | √ | √ | - | 62.37 | 74.52 | 86.49 | 74.46 |
| Flan-T5-large | 780M | √ | √ | - | 67.02 | 76.11 | 87.76 | 76.97 |
| Flan-T5-XXL | 11B | √ | √ | - | 67.81 | 76.32 | 87.98 | 77.37 |
| | | √ | √ | √ | **70.58** | **79.34** | **89.83** | **79.92** |

**Note**: The best results are in **bold** and the second-best are underlined. **ETF** refers to the entity filter model, **AT** refers to the entity typing auxiliary task, and **ICL** represents in-context learning.

As shown in Table **13**, increasing the model size consistently improves performance across all datasets, with larger models demonstrating stronger SciNER capabilities. However, once the model reaches a certain scale, the performance gain becomes marginal. For example, under the same component setting, the performance difference between Flan-T5-large and Flan-T5-XXL is relatively small. After incorporating the ICL module into Flan-T5-XXL, TdSciNER achieves the highest F1 scores of 70.58%, 79.34%, and 89.83% on SciERC, JNLPBA, and BC5CDR, respectively.

**Table 14. Ablation study of different components in TdSciNER across datasets ($F_1$ %).**

| Ablation Setting | SciERC | JNLPBA | BC5CDR |
|---|---|---|---|
| Full TdSciNER | 70.58 | 79.34 | 89.83 |
| *w/o* Entity type filter | 68.32 (-2.26) | 76.53 (-2.81) | 89.42 (-0.41) |
| *w/o* Auxiliary task | 69.59 (-0.99) | 78.69 (-0.65) | 89.47 (-0.36) |
| *w/o* ICL | 67.81 (-2.77) | 76.32 (-3.02) | 87.98 (-1.85) |

To further evaluate the contribution of each component in TdSciNER, we conduct ablation experiments by separately removing the entity type filter, the auxiliary entity typing task, and the ICL selection module, and then evaluating the resulting performance across the three datasets. As shown in Table **14**, removing the entity type filter leads to F1 drops of 2.26%, 2.81%, and 0.41% on SciERC, JNLPBA, and BC5CDR, respectively. This decline indicates that the entity type filter plays an important role in narrowing the candidate type space, thereby reducing incorrect entity assignments by filtering out irrelevant entity types. Additionally, removing the auxiliary entity typing task also degrades performance, although the impact varies across datasets. For complex datasets such as SciERC and JNLPBA, which contain more entity types, the auxiliary task significantly enhances the model's ability to distinguish between entities by providing richer contextual representations. In contrast, since BC5CDR contains only two entity types, the multi-class entity typing task is simplified into a binary classification problem, providing relatively limited benefits for representation learning. Finally, we observe that the contribution of the k-shot ICL strategy increases with the number of entity types. On SciERC and JNLPBA, which contain more complex entity type structures, removing k-shot ICL results in considerable performance drops of 2.77% and 3.02%, respectively. In contrast, BC5CDR, which has fewer entity types, shows a smaller decrease of 1.85%. This trend indicates that as the number of entity types increases, in-context demonstrations become more critical for guiding the model to distinguish among candidate types, thereby improving SciNER performance. Overall, the results observed from the ablation experiments validate that each component make a positive contribution on TdSciNER.

**4.4 Case study**

We conduct a comparative case analysis of our method with SciBERT and GPT-3.5-turbo using illustrative examples from the SciERC dataset, as shown in Table **15**. In these examples, terms such as "machine learning algorithms" and "rule-based learning algorithm" are correctly recognized by all three models. However, both SciBERT and GPT-3.5-turbo fail to detect the term "clauses". This omission may be attributed to the need for deeper contextual understanding and domain-specific knowledge, which these models may not fully capture in this case. In contrast, TdSciNER benefits from domain-specific fine-tuning and the similarity-diversity balanced selection of few-shot demonstrations, enabling it to correctly identify "clauses". This result demonstrates the stronger contextual understanding and robustness of TdSciNER in handling nuanced scientific terminology, further highlighting the value of domain-specific adaptation for SciNER.

**Table 15. Entity recognition comparison of SciBERT, GPT-3.5-turbo, and our TdSciNER on a sample from the SciERC dataset. Entities are color-coded for clarity: cyan for "Generic," purple for "Method," and yellow for "Other (Other-Scientific-Term)."**

| | |
|---|---|
| **Text** | We then use the predicates of such clauses to create a set of domain independent features to annotate an input dataset, and run two different machine learning algorithms: SLIPPER, a rule-based learning algorithm, and TiMBL, a memory-based system. |
| **Labeled** | **Generic:** SLIPPER, TiMBL |

| Entities | Method: machine learning algorithms, rule-based learning algorithm, memory-based system<br>Other: clauses, domain independent features |
|---|---|
| SciBERT | **Task:** predicates<br>**Generic:** SLIPPER, TiMBL<br>**Material:** dataset<br>**Method:** machine learning algorithms, rule-based learning algorithm<br>**Other:** set, annotate |
| GPT-3.5-turbo | **Task:** annotate<br>**Generic:** predicates, features, dataset<br>**Method:** machine learning algorithm, memory-based system<br>**Other:** SLIPPER, TiMBL |
| TdSciNER (**ours**) | **Material:** input dataset<br>**Generic:** SLIPPER, TiMBL<br>**Method:** machine learning algorithms, rule-based learning algorithm<br>**Other:** clauses, features |

# 5. Discussion

This paper presents TdSciNER, a model designed to leverage entity type information to enhance LLM performance in SciNER tasks. In this section, we discuss the broader implications of our findings and outline the limitations of the study.

### 5.1. Implications

Recognizing entities in scientific papers is a fundamental task in scientific text mining and is crucial for downstream applications such as knowledge discovery, document retrieval, and question-answering systems. In recent years, the use of advanced generative LLMs for SciNER has attracted increasing attention. However, as a domain-specific NER subtask, SciNER involves more complex and fine-grained entity types. Moreover, prior work that reformulates sequence labeling as a generative task has not fully exploited entity type information to improve SciNER performance. To address this gap, we introduce TdSciNER, an entity type-driven SciNER model that achieves competitive results comparable to supervised learning baselines across three publicly available scientific datasets. In our model, we first construct an entity type filter to exclude irrelevant types, which helps LLMs focus on entity types that are more likely to appear in the input text and significantly improves SciNER performance. This result is consistent with the observations of Wang et al. (2022b), who indicated that reducing the entity type options in prompts enables LLMs to make more accurate NER predictions. Second, during the fine-tuning stage, we design an auxiliary multi-class entity typing task to enrich semantic representations without additional labeling costs. Ablation experiments show that multi-task learning also enhances entity recognition performance in scientific domains. These findings are consistent with earlier studies on the effectiveness of auxiliary tasks in information extraction (Martins et al., 2019; Wang et al., 2021a; Yang & Mitchell, 2016). Finally, in low-resource scenarios, utilizing few-shot ICL to guide LLMs in completing NER tasks proves to be a universal approach (Wang et al., 2022b; Wang et al., 2023b). Our results further reveal that selecting demonstrations with high sentence similarity, diverse entity types, and varied sentence structures enables the model to better capture domain-specific entity characteristics, thereby improving SciNER accuracy.

It is worth noting that TdSciNER also shows strong potential for cross-domain SciNER applications. In interdisciplinary fields such as bioinformatics, scientific entities often involve more diverse and domain-specific characteristics, making accurate recognition particularly challenging. In this context, entity type information provides an effective signal for helping LLMs distinguish among different entity types and achieve more accurate recognition. By incorporating domain-specific knowledge into the SciNER process, TdSciNER not only improves entity recognition accuracy but also supports downstream applications such as knowledge extraction and information retrieval in scientific contexts.

More broadly, the effectiveness of incorporating domain-specific knowledge into LLMs has recently been demonstrated in a variety of specialized domains beyond scientific text mining. Recent studies in engineering and industrial intelligence have shown that integrating expert knowledge, structured knowledge representations, and human feedback with LLMs can substantially improve task performance, interpretability, and reliability (Zhang et al., 2025; Gao et al., 2026). For instance, DiagLLM (Wang et al., 2025b) combines expert knowledge and multimodal information with LLM-based reasoning to achieve explainable bearing fault diagnosis, demonstrating enhanced diagnostic accuracy, interpretability, and cross-domain generalization. These developments suggest that aligning LLMs with structured domain knowledge is a promising direction for enhancing the reliability, explainability, and generalization capability of LLMs in specialized applications. Our findings provide further evidence that type-driven knowledge can serve as an effective mechanism for adapting LLMs to domain-specific tasks.

### 5.2 Limitations

We acknowledge several limitations of this study. First, TdSciNER does not yet surpass highly specialized pre-trained models such as SciBERT and BioBERT on certain datasets. This is largely because these models are pre-trained on large-scale domain-specific corpora, enabling them to capture fine-grained terminology and contextual patterns that may not be fully learned by general-purpose LLMs, even after fine-tuning. Second, our experiments are conducted mainly on datasets from scientific domains. Future work could evaluate TdSciNER on a broader range of domain-specific datasets, such as those from law or finance, to further assess its robustness and generalizability across different fields. Third, the prompt design may introduce a certain degree of subjectivity, which could influence model performance. Therefore, exploring more systematic and automated prompt generation methods would be a promising direction for reducing potential bias caused by manual intervention. Fourth, the auxiliary task in this study primarily focuses on entity type classification. Incorporating other information extraction tasks, such as relation extraction, may provide richer contextual signals and further improve the effectiveness of the main SciNER task. Finally, we acknowledge the potential risk of data contamination, as the datasets used in this study may have been included during the pre-training of T5 or the instruction tuning of Flan-T5. However, the primary goal of our experiments is to compare the proposed method with baseline approaches under the same evaluation setting. Since all methods are built upon comparable pre-trained models, such as BERT, SciBERT, Flan-T5, or GPT-family models, and are evaluated on the same datasets using the same evaluation metrics, data contamination is unlikely to substantially affect the relative performance comparison.

# 6. Conclusion and future work

In this paper, we introduce TdSciNER, a novel type-driven multi-task learning framework that leverages entity type information to improve SciNER performance. The framework consists of three key components. First, we introduce a lightweight entity type filter model to identify potential entity types in scientific texts. Second, we design an entity typing task as an auxiliary task and integrate it with the primary SciNER task through multi-task learning to obtain richer semantic representations. Third, we develop a sample selection strategy to choose appropriate labeled examples as prompt demonstrations, thereby activating the in-context learning capabilities of LLMs.

Experiments conducted on three widely used scientific datasets demonstrate that TdSciNER achieves performance comparable to fully supervised baselines. Furthermore, we validate that all three components constructed based on entity type information contribute to the improvement of TdSciNER. First, in SciNER tasks involving multiple entity types, entity type information helps identify the most likely entity types in a sentence, enabling LLMs to narrow the candidate type space and achieve more accurate NER results. Second, we observe that designing entity typing as an auxiliary multi-class classification task allows the model to learn richer textual representations, thereby enhancing its learning ability without requiring additional data annotation. Third, when activating the ICL capabilities of LLMs, both the similarity between demonstrations and the input text and the diversity of demonstration structures contribute to selecting more effective examples.

This work provides valuable insights for applying LLMs to SciNER, as well as to broader scientific information extraction and information retrieval tasks. Meanwhile, we recognize that there remains room for further optimization in model design and effective demonstration selection. In future work, we will explore the integration of more advanced LLMs with domain-specific knowledge to further enhance SciNER performance. Moreover, developing a unified SciNER framework that can be applied across disciplines is also an important direction for future research. Finally, extending scientific entity recognition to a wider range of downstream tasks in scientific text mining may provide meaningful support for the academic community.

# Acknowledgments

This paper was supported by the National Natural Science Foundation of China (Grant No.72074113, 72374103).

# Author conflict statement

The authors declare that there are no conflicts of interest.

# Appendix

In this section, we discuss the prompt construction process used in this work. Our prompt design was inspired by a review of prior studies in LLM-based NER, including work such as GPT-NER (Wang et al., 2023b), InstructUIE (Wang et al., 2023c) and SLIMER (Zamai et al., 2024), and other instruction-tuned approaches, from which we extracted common practices and effective strategies. Based on this analysis, we designed our initial prompt template to incorporate the following key elements, along with an explanation of their intended roles in guiding the LLM:

1、 **Task Instruction**: A concise natural language description specifying the task objective (e.g., “Please list all scientific entities of [*C*] in the following text.”). This helps the LLM understand the task requirements and generate accurate target outputs.

2、 **Entity Type Specification**: A clearly defined list of target entity types, tailored to each dataset (e.g., *Task*, *Method*, *Material*, *Metric*, *Generic*, *Others*). This component narrows the model’s focus to only relevant types, preventing irrelevant or overly generic entity predictions, and improving both precision and recall.

3、 **Output Format Constraint**: Instructions on how the model should present its output (e.g., as a JSON list of entities or a structured text format). We standardize the output format to facilitate automatic parsing and evaluation, and to reduce the risk of ill-formed or inconsistent responses that may occur with open-ended generation.

4、 **Demonstration Examples** (optional): This component provides a few annotated examples to activate the ICL capability of the LLM to better understand the task and generate outputs in the expected format. In our work, we include such examples to offer informative and representative cues (see Section 3.4 for details).

Based on these elements, we first designed a standard prompt for our TdSciNER, which consists of **a task instruction, a list of entity types, and output format requirements, a**s shown in Figure **3**. To provide the model with additional contextual cues, we append a **Note** at the end of the prompt indicating that entity types in the subset [C′] may exist in the input text. Here, [C′] is a high-probability subset of the full type set [C], computed based on the fine-tuned entity type filter model (Section 3.2). While the prompt explicitly asks the model to extract entities of a specific type from [C], the inclusion of [C′] reinforces type-level signals that are more likely to appear in the context and provides soft guidance on the expected type distribution within the sentence.

**Scientific named entities recogenition with LLMs**

**Input**: Please list all scientific entities of type [*C*] in the following text. Your output should follow the JSON format: {'entity type': [entities]}. If no entities, return None.
**Text**: $x$
**Note**: Entities of type [*C′*] may exist in the text.

**Figure 3. The prompt for LLMs implementing SciNER based on entity type information.**

To control for stylistic variability, we maintained consistent prompt templates across all datasets and models. Accordingly, we further designed the prompt for the entity typing task in a similar style, as illustrated in Figure **4**. The prompt for the entity typing task follows the same structure as the SciNER task (Figure **3**) but differs in objective: instead of extracting entities from text, the model is required to perform a multi-class classification by assigning each given entity to one of the predefined types. Since both tasks operate on the same datasets and share the same type inventory, no additional labeling effort is required.

**Entity typing as an auxiliary task for SciNER**

**Text**: $x$
**Input**: Please typing these scientific words [entities list] from the text. Type options are [*C*]. Your output should follow the JSON format:{'entity type': [entities]}.

**Figure 4. The prompt for the entity typing auxiliary task.**

Finally, building upon the design in Figure **3**, we further incorporated few-shot demonstrations into the prompt to better activate the in-context learning capabilities of the model, as illustrated in Figure **5**. The criteria and procedure details for demonstration selection can be found in Section 3.4. Please note that this figure uses the SciERC dataset as an example, and for all other datasets, both the entity types and few-shot examples are selected from the corresponding training sets to prevent potential data leakage.

**Enhancing scientific entity recognition with Shot-ICL**

**Input**: Please list all scientific entities of type [Task, Method, Metric, Dataset] in the following text. Your output should follow the JSON format: {'entity type': [entities]. If no entities, return "None".
Here is a example:
**Text**: This model improved the performance in image classification task.
**Output**: {"Task": ["image classification"]}
**Text**: $x$
**Note**: Entities of type [$C'$] may exist in the text.
**Output**:

**Figure 5. Prompt design for enhancing scientific named entity recognition with *k*-Shot in-context learning (example from the SciERC dataset).**

In summary, our prompt design aligns closely with prior work on LLM-based NER, allowing for fair and meaningful comparisons. To ensure consistency across experiments, we applied the same prompt structure and stylistic format to all models and datasets, which helps minimize variability introduced by prompt design choices. We acknowledge concerns that alternative prompt formulations might lead to different outcomes. However, prompt engineering is not the main focus of this paper. To encourage further exploration in this direction, we publicly release our code and data at https://github.com/tongbao96/code-for-SciNER, encouraging the community to investigate the impact of different prompt strategies and advance research in LLM-based SciNER.